\documentclass[runningheads]{llncs}
\usepackage[T1]{fontenc}
\usepackage{graphicx}

\usepackage{cite}
\usepackage{amsmath,amssymb,amsfonts}
\usepackage{algorithmic}
\usepackage{graphicx}
\usepackage{textcomp}
\usepackage{xcolor}
\usepackage{tabularx}
\usepackage{url}
\usepackage{csquotes}

\begin{document}
\title{Enhance the after-discharge mortality rate prediction via learning from the medical notes} 
\titlerunning{Enhance the after-discharge mortality rate prediction}
%
\author{Zijiang YANG \inst{1} }
\authorrunning{Zijiang YANG}
%

\institute{University of Texas at Austin, Austin TX 78712, USA
\email{zy4957@eid.utexas.edu} 
}
\maketitle              

\begin{abstract} 
With the increase of the Electronic Health Records (EHR) data, more and more researchers are developing machine learning models to learn from the medical notes. These unstructured text data pose significant challenges on the learning process as the quality of data is low. These data are often messy, repetitive and redundant. We have shown these notes data to be informative by conducting the after-discharge mortality rate prediction task. The AUC-ROC for models using the medical note information is generally 0.1 higher than those without the medical notes. Furthermore, we propose the Deep Neural Network(DNN) model with ‘pooling' mechanism to enhance the mortality prediction. Based on the experimental results, we demonstrate that the proposed model outperforms the traditional machine learning models like the tree-based models. The proposed method learns from the most informative medical notes and improves the prediction accuracy significantly. The AUC-ROC for the proposed model is 2\% to 14\% higher than the traditional ones in 15-days, 30-days, 60-days, 365-days after-discharge mortality prediction tasks. Moreover, we can discover some interesting knowledge through the traditional and proposed models. These knowledge are inspiring but also consistent with the previous findings. The models are able to reveal the relationships between the informative keywords and documents from the medical notes and the severity of the patients.

\end{abstract}

\section{Introduction}
The Electronic Health Records (EHR) data has revolutionized the patient management, treatment planning, readmission and length-of-stay prediction. The data consist of both the basic information and detailed diagnoses and treatments, providing unprecedented information to discover the patients' health condition and study the disease progression. However, the data contains both the structured and unstructured ones. Structured data such as the gender, age, ethnicity and marital status are easy to process and analyzed. Researchers are currently focusing on harness the power from the unstructured data such as the medical notes. However, these unstructured text data pose significant challenges on the learning process as the quality of data is low. These data are often messy, repetitive and redundant.\cite{Weir2007} 

Although the unstructured data is of low quality, to show the usefulness of these notes data, we conduct the experiments on the after-discharge mortality rate prediction task. Specifically, we conduct the experiments on the patients diagnosed as \enquote*{kidney failure}\cite{Francis2024, Kim2019}. Kidney failure, also known as renal failure, means one or both of your kidneys no longer function well. It is sometimes temporary and develops quickly, which is characterized as  acute kidney failure. Or it may be chronic that develops over time. Predicting the mortality for kidney failure patients is crucial to understanding the progression of such disease. Previously, researchers develop traditional models using basic information only. However, the medical notes such as nursing notes, radiology and discharge notes may contain much predictive information. Actually, it is reported that the AUC-ROC for models using the medical note information is generally 0.1 higher than those without the medical notes based on our experiments.

Furthermore, we propose the implementation of a Deep Neural Network (DNN) model that integrates an advanced \enquote*{pooling} mechanism to enhance the accuracy and effectiveness of mortality prediction. This pooling mechanism, which involves weighting and aggregating information across different categories of the medical notes. It is improving the model’s ability to by capturing the important information in the medical notes data. By leveraging this mechanism, the DNN model is able to retain important information and ignore the redundant one, thus making the model more predictive.

Through comprehensive experiments and evaluation, we demonstrate that our proposed DNN model significantly outperforms traditional machine learning models, particularly tree-based models such as  Random Forests, and Extreme Gradient Boosting model. The experimental results show that the proposed DNN model consistently achieves higher performance metrics in mortality prediction across various time span. Specifically, the Area Under the Curve of the Receiver Operating Characteristic (AUC-ROC) for the DNN model is observed to be between 2\% and 14\% higher than that of the traditional tree-based models. 

Lastly, we can discover new knowledge based on the models. We are able to see why the inclusion of the natural language information improves the model significantly. What medical keywords are effective in predicting the mortality and understanding the disease. Also, we can discover the importance of different medical notes through the proposed model. Interestingly enough, these knowledge are consistent with the previous findings. For example, we discover that the medical notes under 'Discharge summary' category are most informative. 

The paper is organized as follows: section II introduces the related work; section III gives the details of the methodology; section IV presents and analyze the experimental results; section V shows the discovered knowledge from the traditional and the proposed model; section VI concludes this paper.

\section{Related work}
Previous researchers have done a lot of work on analyzing the electronic health records(EHR) and solving multiple tasks. However, we still have much difficulty in understanding and extracting the useful information from the vast amount of EHR data.

Weir et al.\cite{Weir2007} identified difficulties in using the electronic medical documentation: 1) information overload; 2) hidden information; 3) lack of trust; These factors will be detrimental to making decisions.

Edin et al.\cite{Edin2023} reproduced, compared, and analyzed state-of-the-art automated medical coding machine learning models. They noted the weakness of previous work due to weak configurations, poorly sampled train-test splits, and insufficient evaluation. 

Shanafelt TD et al.\cite{West2018} and Grover et al.\cite{Grover2018} pointed out that the physicians would burn out and blindly copy the computer generated data to the medical notes.  These will inevitably create much unnecessary and irrelevant information that may not reflect the patients' symptoms. 

Despite the difficulty in extracting the information from the EHR data, previous researchers have tried various techniques such as gradient boosting\cite{AEW2017}, random forests\cite{Awad2017}, support vector machines\cite{Ghassemi2014}, ensemble methods\cite{Awad2017} and stacking method\cite{Rashidy2020}.  Most of the researchers research on the structured data such as the admission information and  physiological measurements. Very few of them are using the information from the medical notes\cite{IER2018, Kocbek2017, Sushil2018, Zalewski2017}. Many researchers have done work on in-hospital mortality prediction\cite{Ghassemi2014, Sushil2018, Tran2018}. There is limited work in the field of predicting the after-discharge mortality\cite{Kocbek2017}.

\section{Methodology}
\subsection{Basics of the data}
The study uses MIMIC III\cite{Johnson2016} data to learn medical notes. MIMIC-III is a large and freely available database comprising deidentified health-related data associated with more than 40,000 patients who remained in critical care units of the Beth Israel Deaconess Medical Center between 2001 and 2012.  The MIMIC-III Clinical Database is available on PhysioNet. \url{https://physionet.org/content/mimiciii/1.4/}

In particular, we used the following tables from the dataset: 
\begin{itemize}
    \item \textit{patient}
    This records the basic information of the patient such as the age, ethnicity, marital status and so on.
    \item \textit{admission}
    This records the ICU admission and discharge information for the patients.
    \item \textit{diagnoses} This records the diagnoses for the patients.
    \item \textit{ICD diagnosis} This records the ICD code of the diagnosis for the patients.
    \item \textit{noteevents }   This records the medical notes for the patients.
\end{itemize}

\subsection{Pre-processing of the data}

In our experiments, for illustration, we focus on the patients that are diagnosed as ‘kidney failure'. We filter out the outliers of patients with age over 120 and the patients with ‘DEAD/EXPIRED' discharge location. We are left with 6365 legitimate patients.
The target label we want to prediction is the mortality after discharge from the ICU. Specifically, based on the discharge date and the death date, we compute the label ‘Survive 30 days after discharge' for each patient: 1 for survival and 0 for death. The same process is repeated for 15-days, 60-days and 365-days.

\begin{table}[]
\centering
\begin{tabular}{c c }
\hline
ADMISSION TYPE & ADMISSION LOCATION \\ \hline
INSURANCE      & LANGUAGE           \\ \hline
RELIGION       & MARITAL STATUS    \\ \hline
GENDER         & ADMIT AGE           \\ \hline  

\\ 
\end{tabular} 
\caption{  Basic information of the admitted patients}
\label{tab:basicinfo}
\end{table}

\subsubsection{Basics of admission information}
The admission table joining the patient table provides the basics admission information. Such information includes 8 features such as ADMISSION TYPE, ADMISSION LOCATION, INSURANCE and so on. All the information is tabulated in Table.\ref{tab:basicinfo}. 7 features are categorical features and ADMIT AGE is a continuous feature. The categorical features are transformed into 0/1 features and continuous features are set to be float.

\subsubsection{Statistics of the medical notes}

We first gather all medical notes corresponding to 6365 patients and apply the tokenization technique from the \textit{nltk} package to tokenize each medical note. The basic statistics for the medical notes are displayed in the table.\ref{tab:notecatogory}.  Majority of the medical notes are Nursing/other notes followed by the Radiology reports. These two consist of more than half of the medical notes. The discharge summaries have the longest token length among all the categories as they are typically very long.

After the tokenization, the \textit{CountVectorizer} in \textit{sklearn} package is used to count the frequency for the top 400 most frequent tokens. The counts  of the tokens form a 400-dimensional feature vector. For traditional methods mentioned below, we count the number of tokens of all the medical notes for each hospital admission ID(HADM\_ID).

\begin{table}[]
    \centering    
    \begin{tabular}{@{}lrlr@{}}
    \hline
    Category & Count                                  & Percentage                     & Average token length \\
    \hline
    
    Nursing/other                          & 80644                          & 0.324526             & 153.2393            \\ 
    Radiology                              & 55368                          & 0.222811             & 185.6273            \\
    Nursing                                & 39717                          & 0.159828             & 264.2057            \\ 
    Physician                              & 26705                          & 0.107466             & 751.9017            \\
    ECG                                    & 23053                          & 0.092769             & 30.2879            \\ 
    Discharge summary                      & 7396                           & 0.029763             & 1623.0770            \\
    Respiratory                            & 5917                           & 0.023811             & 145.7208            \\ 
    Echo                                   & 4983                           & 0.020052             & 320.1120             \\
    Nutrition                              & 1794                           & 0.007219             & 268.7352            \\ 
    General                                & 1487                           & 0.005984             & 204.1432            \\
    Rehab Services                         & 806                            & 0.003243             & 403.7506            \\ 
    Social Work                            & 342                            & 0.001376             & 306.4766            \\
    Case Management                        & 239                            & 0.000962             & 144.0084            \\ 
    Pharmacy                               & 27                             & 0.000109             & 229.4815            \\
    Consult                                & 20                             & 0.000080             & 880.5000              \\
    \hline
    \\
    \end{tabular}
    \caption{Categories of the medical notes}
    \label{tab:notecatogory}
\end{table}

\begin{figure}
    \centering
    \includegraphics[width=1.0\linewidth]{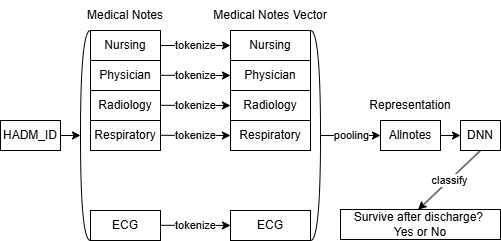}
    \caption{Flow chart for medical notes representation}
    \label{fig:Flowchart}
\end{figure}


\subsection{Traditional models}
This mortality prediction task is basically a binary classification task. For comparison, we set up the traditional
machine learning models to perform the classification. Followings are the model we consider:
\begin{itemize}
    \item \textit{Logistic regression} The logistic regression model predicts probabilities using a sigmoid function (also called the logistic function), which maps any input value (a linear combination of input features) to a value between 0 and 1.
    \item \textit{Random forest}  This is an ensemble learning method used for both 
classification and regression tasks. It combines multiple decision 
trees to improve the overall model's accuracy and robustness.
    \item  \textit{XGBoost} This model, Extreme Gradient Boosting, is an optimized, scalable, and 
efficient implementation of Gradient Boosting. It is widely used in 
machine learning tasks, especially for structured/tabular data.
\end{itemize}

The above models achieves decent results in the traditional downstream tasks like readmission rate prediction and length-of-stay prediction\cite{Edin2023}.

\subsection{Proposed method}
The proposed method uses the 'pooling' mechanism to gather all the feature vectors. For each category, we count the tokens and form a 400-dimension feature vector $f_i$. The representation for all notes, denoted as $f_{allnotes}$, is computed as follows:

\begin{equation}
    f_{allnotes} = \sum^{N}_{i=1} f_iw_i
\end{equation}

where $w_i$ is weight for $i$-th category and $N$ is number of categories.
Previous researchers treat all text documents equally, where $w_i$ is the same for all $i$. It is inherently the idea of the Deep Average Network (DAN) \cite{iyyer2015}. However, medical notes often contain repetitive content. For example, 
 current Nursing notes often repeats the content of the previous Nursing note. In addition, different types of medical notes convey different messages. Treating them equally will down-weigh the important ones and up-weigh the 
 repetitive ones. 

We build up our method by constructing the pooling weight and passing it into a Deep Neural Network(DNN) as a learnable parameter. In our experimental settings, we configure the DNN as 4-layer neural network with 2 hidden layers. The hidden layer size is set to be 70. All the hidden layers have short-cut links interconnected.

To train the model, we carefully initialize the weights, use the Adam optimizer and set the learning rate to be as  small as 0.0005. As the over-fitting issues are fairly common in medical tasks, we use the validation set to early-stop the training process.

\section{Experimental results}
\begin{figure}
    \centering
    \includegraphics[width=.75\linewidth]{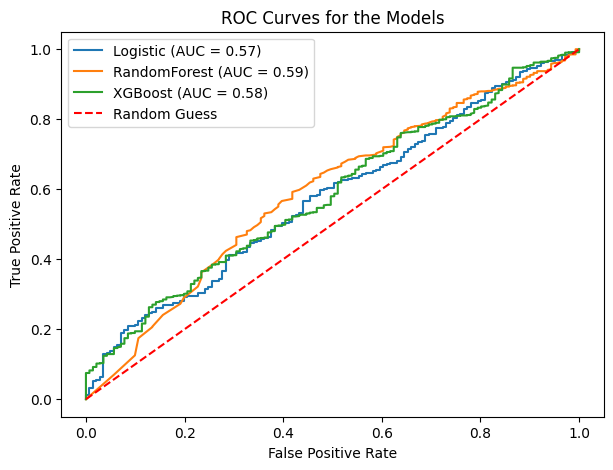}
    \caption{AUC-ROC curve on 30-days mortality prediction using multiple machine learning classifiers without medical notes}
    \label{fig:rocnonotes30}
\end{figure}

We first start with the simplest mortality prediction based on the basic patient information only. The survival 30 days after discharged is predicted using the traditional methods mentioned above. The samples are randomly split into the training, validation and test sets with ratio $7:1.5:1.5$. The hyper-parameters are learned via the validation set. The experimental results are reported as the AUC-ROC graph in the Figure.\ref{fig:rocnonotes30}.

\subsection{Baseline model using patient basic information only}

 It is reported that these 3  models have very similar model accuracy. Based on the result, we can see that these models cannot deliver  a good result. The AUC is just around 0.58, which barely outperforms the random guessing.

\subsection{Baseline model using medical notes information}

By tokenizing the medical notes and constructing them as input feature vectors, we repeat the experiments using the same models. We are able to observe a significant improvement: all models perform better than basic information only; the AUC increase by 0.09 from 0.59 to 0.68 for the Random forest method. This results  demonstrate the effectiveness of the medical notes in mortality prediction, which contradicts to the Ghassemi et al.\cite{Ghassemi2014} as they used a limited set of structured variables.

\begin{figure}
    \centering
    \includegraphics[width=0.75\linewidth]{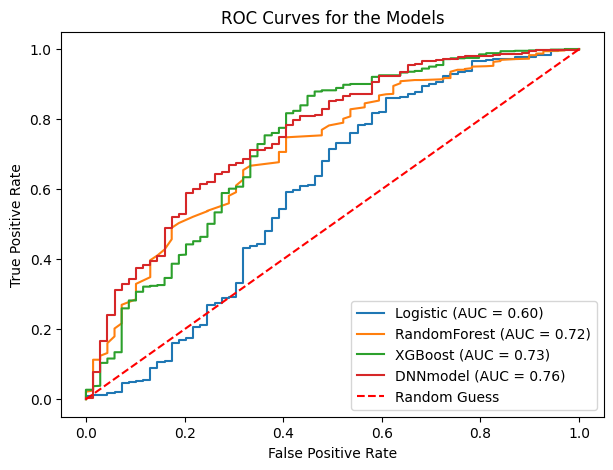}
    \caption{AUC-ROC curve on 15-days mortality prediction using multiple machine learning classifier with medical notes}
    \label{fig:enter-label} 
    \centering
    \includegraphics[width=0.75\linewidth]{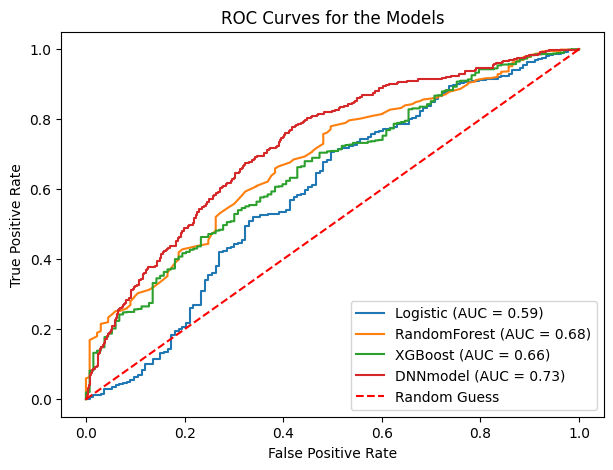}
    \caption{AUC-ROC curve on 30-days mortality prediction using multiple machine learning classifier with medical notes}
    \label{fig:enter-label}




    \centering
    \includegraphics[width=0.75\linewidth]{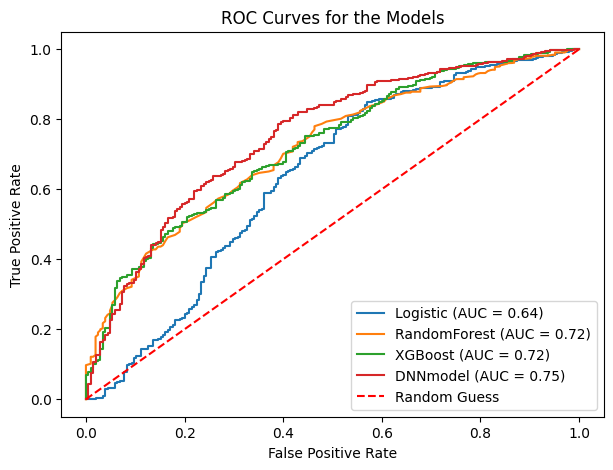}
    \caption{AUC-ROC curve on 60-days mortality prediction using multiple machine learning classifier with medical notes}
    \label{fig:enter-label} 
    
    \centering
    \includegraphics[width=0.75\linewidth]{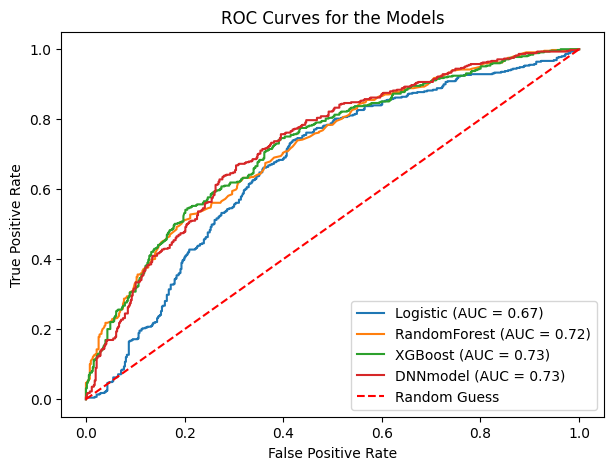}
    \caption{AUC-ROC curve on 365-days mortality prediction using multiple machine learning classifier with medical notes}
    \label{fig:enter-label}
\end{figure}

We then apply our method on the mortality prediction task for 15-days, 30-days, 60-days and 365-days time spans. It is reported that this DNN based method achieves best performance across all the time spans. It outperforms the traditional machine learning methods by 2\% to 14\% based on the AUC-ROC criterion. It is worth noting that the DNN gives a better performance than the prevailing tree based methods for the medical prediction task.

\begin{table} [t]
\centering
\begin{tabular} {l| r| r| r| r}

    \hline
Category          &   Weight   & Sensitivity   & Normalized   & Token Length \\ 
  &     &   &   Sensitivity   &   Correlation with Survival\\  \hline  
  
Nursing/other     & \textbf{0.255617} & 0.002279    &   0.349164              & -0.02560                         \\
Radiology         & 0.036422 & -0.000036    & -0.006710             & \textbf{-0.11285}                        \\
Nursing           & -0.163920 & 0.001789    &   0.472738              & -0.07814                        \\
Physician         & 0.001094 & 0.000506    &   0.380522              & -0.08529                        \\
ECG               & 0.020193 & -0.002660    &  -0.080600             & -0.03462                        \\
Discharge summary &\textbf{ 0.243545} & -0.001050    &  \textbf{-1.697190}            & \textbf{-0.22884}                       \\
Respiratory       & \textbf{0.230256} & -0.000340    &  -0.049690              & -0.05800                          \\
Echo              & 0.082943 & -0.002190    &  -0.699590              & -0.07606                        \\
Nutrition         & -0.185750 & -0.001080    & -0.290890              & -0.06459                        \\
General           & 0.150531 & 0.000165    & 0.033651              & -0.04432                        \\
Rehab Services    & -0.082580 & 0.004298    & 1.735236              & -0.04160                         \\
Social Work       & 0.058556 & 0.001955    & 0.599275              & -0.03227                        \\
Case Management   & -0.192240 & -0.006180    & -0.889630             & -0.05788                        \\
Pharmacy          & -0.089710 & 0.000913    & 0.209617              & -0.01427                        \\
Consult           & -0.121180 & -0.000430    & -0.374790              & -0.00873                        \\

\hline 
\end{tabular}
\caption{Sensitivity analysis of different categories of medical notes (Significant values are bolded)}
\label{tab:categoryweight}
\end{table}


\begin{figure}
    \centering
    \includegraphics[width=1\linewidth]{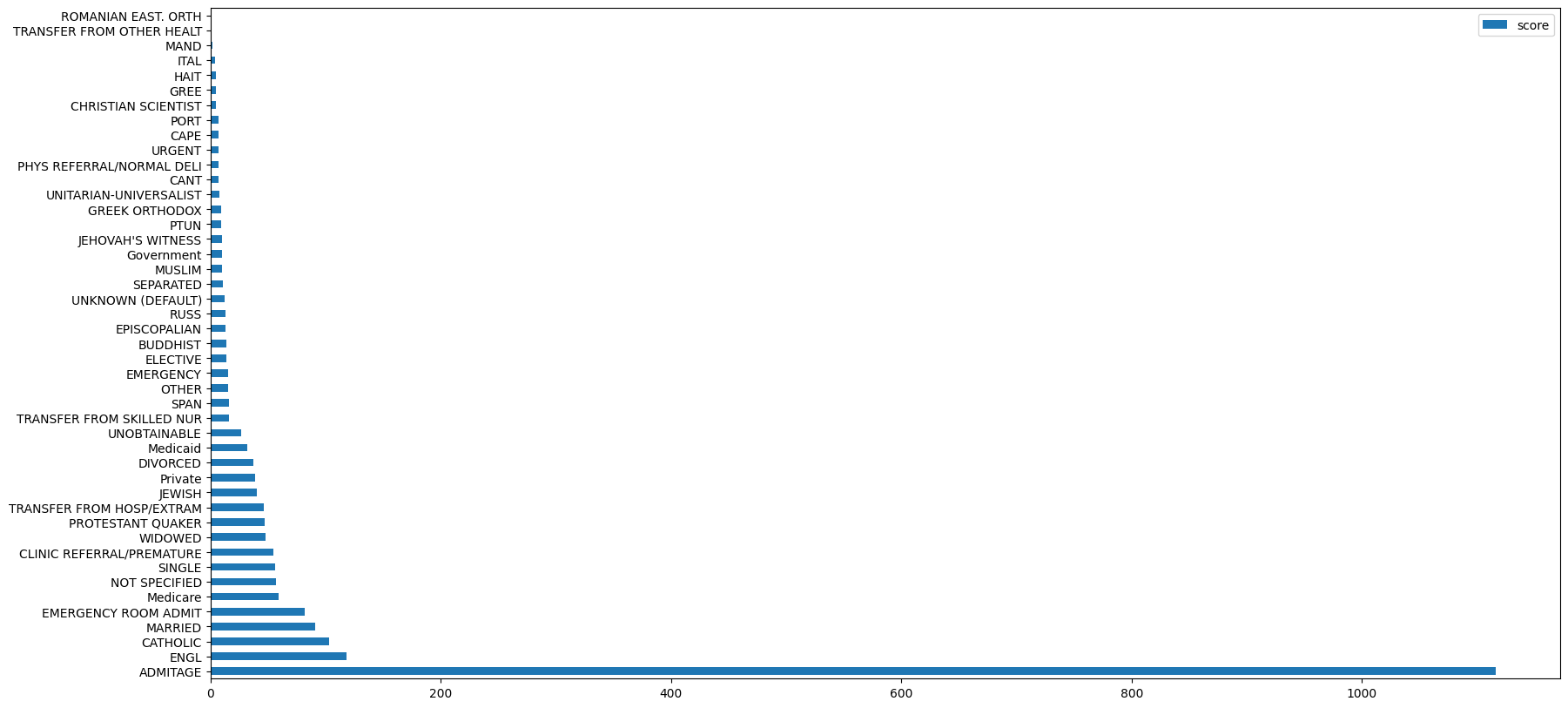}
    \caption{XGBoost feature importance on 30-days mortality prediction using  basic information only}
    \label{fig:featureimportance} 
    
    \centering
    \includegraphics[width=1\linewidth]{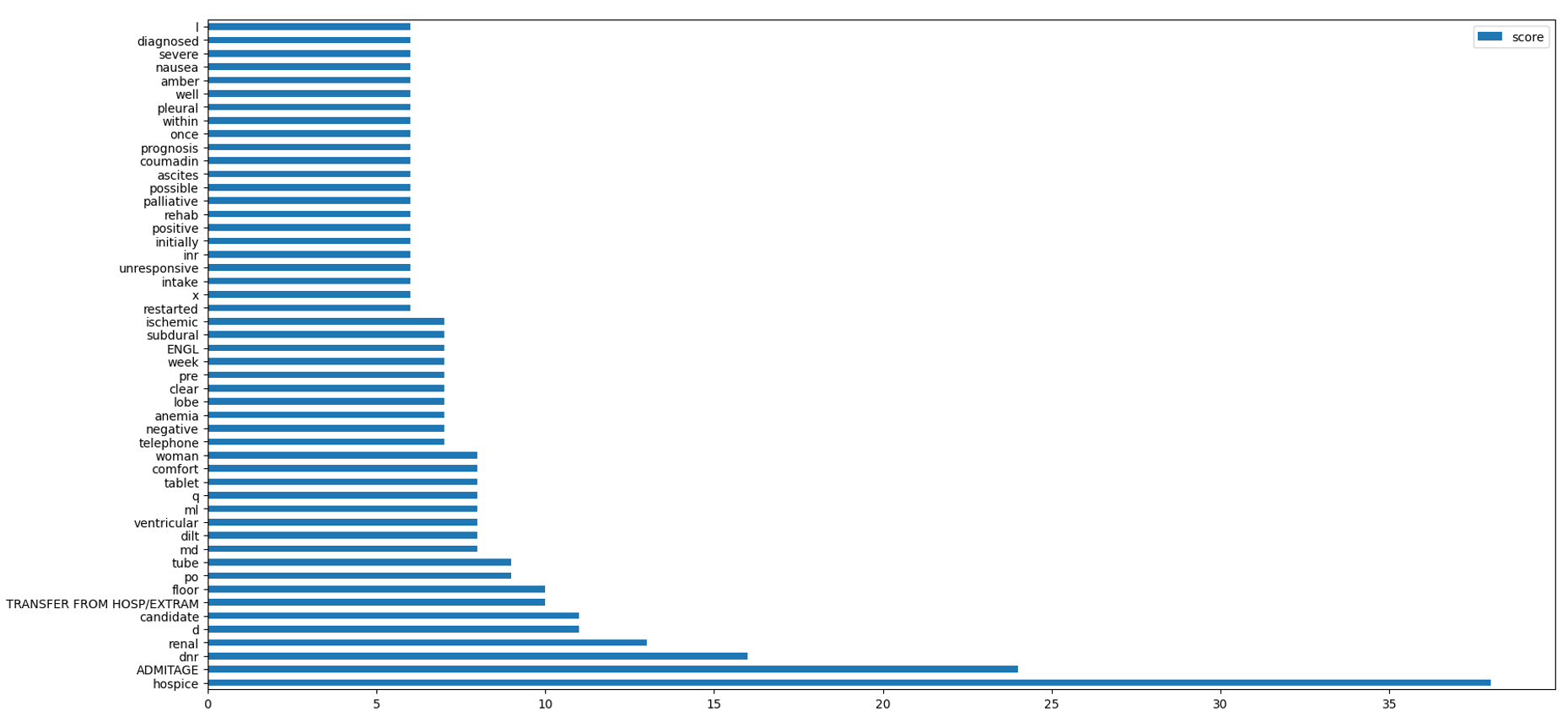}
    \caption{XGBoost feature importance on 30-days mortality prediction using  basic information and medical notes}
    \label{fig:featureimportance_keywords}
\end{figure}

\section{Knowledge discovery}
\subsection{Knowledge from traditional methods}
We conduct the analysis of the feature importance for the traditinal model XGBoost. The result is shown in the Figure.\ref{fig:featureimportance}. It is shown that the ‘ADMIT AGE’ admission age, ‘ENGL’ for English language speaker, ‘CATHOLIC’ for catholic religion are the three most important features. In particular, the continuous feature ‘ADMIT AGE’ admission age has a dominant impact on the prediction accuracy. Its importance is much larger than the other features as is shown in the figure.

Then we conduct the analysis of the feature importance for the XGBoost using both the basic information and the medical notes. The result is shown in the Figure.\ref{fig:featureimportance_keywords}. In this scenario, it is shown that the ‘hospice’, ‘ADMIT AGE’ for admission age, ‘dnr' for Do-Not-Resuscitate are the three most important features. Here, we see that the feature ‘ADMIT AGE’ admission age is no longer a dominant impact. Other medical keyword features, extracted based on the natural language processing analysis from the medical notes, come into the play. The keyword features contribute significantly to the model prediction.

For example, we have the keyword ‘hospice' listed as the most important feature. Hospice  care is a service for people with serious illnesses who choose not to get (or continue) treatment to cure or control their illness. According to the Figure.\ref{fig:featureimportance_keywords}, it is a very strong indicator for the mortality of the kidney failure patients. However, does it mean that the patients with ‘hospice’ key word in their note event will definitely die soon? Actually no. In fact, it just indicates the patients may be under severe condition. 36.67\% of the patients with ‘hospice’ will survive in the 30 days after the ICU and 10.25\% of the patients will survive in the 365 days. Also, the patients with 'dnr' are also under severe condition. A DNR order is a legal document that means a person has decided not to have Cardiopulmonary resuscitation (CPR) attempted on them if their heart or breathing stops. These important medical keywords learned from the machine learning models shed the light on the underline relationship between the patients’ illness condition and their mortality rate.

\subsection{Knowledge from proposed DNN method}
Based on the DNN model constructed, we are able to perform the analysis on the medical notes. Firstly, we can analyze the weights of each category passed into the DNN model. The table.\ref{tab:categoryweight} tabulates the weights. 
As is shown in the table, the categories \enquote*{Nursing/other}, \enquote*{Nursing}, \enquote*{Discharge summary} and \enquote*{Respiratory} contributes most to the prediction result. This is consistent with the preliminary results from Hsu et al.\cite{Hsu2020}. In their experiments on readmission prediction, the discharge summary is the most informative category among the medical notes.

We then compute the sensitivity of the prediction score $y_{pred}$ with respect to each token via:
\begin{equation}
\begin{split}
    \frac{\partial y_{pred}}{\partial countoftoken_i} \approx \frac{\Delta y_{pred}}{\Delta countoftoken_i} \\
=  y_{pred}|{(token_i + 1) -  y_{pred}|token_i }
\end{split} 
\end{equation}

We artificially alter the count of token \textit{i} and see the change in the prediction score $y_{pred}$.
The results are tabulated in the Table.\ref{tab:categoryweight}. As different medical notes have different lengths, we account for this by multiplying the average token length in the Table.\ref{tab:notecatogory}. The normalized sensitivities are generated. The meaning of this quantity is to measure how the prediction score will change when a new medical note is presented. Again, it is reported that the discharge summary is the most influential medical note for predicting the mortality. (Ignore the last few rows as they contains very limited samples.)

Finally, we compute the correlation of the token length and the patients' survival in the Table.\ref{tab:categoryweight}. So, we are investigating how the token length of the medical notes will after survival. Generally, they are negatively correlated with survival, which means longer the notes, more likely the patient will not survive. The discharge summary is most negative one, suggesting that it is a very strong predictor for the survival.

There are also few interesting observations from the Table. It is shown that the radiology and ECG reports are not informative for prediction based on the weight and sensitivity analysis. Physician notes help with the prediction but not as strong as the nursing notes. Lastly, the Echo notes information is also helpful for this prediction task.



\section{Conclusion}
Based on the experiments on the Electronic Health Records (EHR), especially on the disease \enquote*{kidney failure}, our results confirm the value of medical notes for the after-discharge mortality  prediction. The AUC-ROC for the models using medical note information is generally 0.1 higher than that without using the medical notes. The Natural Language Processing techniques are shown to be effective in extracting out the useful information. Moreover, we are also able to discover what keywords in the medical notes would be most informative for understanding the disease condition and its progression.

In addition, we further enhance the mortality prediction by introducing the \enquote*{pooling} mechanism. Through assigning the weights to different categories and learning the weights through the training process, we are able to discover the significance of each medical notes. The performance of our DNN model achieves the best prediction performance for 15-days, 30-days, 60-days and 365-days after-discharge mortality. The proposed model outperforms the traditional models by 2\% to 14\% based on the AUC-ROC criterion. In addition, it is found that \enquote*{Discharge summary} and \enquote*{Nursing} are the most informative categories for predicting the mortality among the categories of the notes.

\end{document}